\documentclass[times,twocolumn,final,authoryear]{elsarticle}
\usepackage{ycviu}
\usepackage{framed,multirow}

\usepackage[colorlinks,
linkcolor=red,
anchorcolor=blue,
citecolor=green
]{hyperref}

\usepackage{amssymb}
\usepackage{latexsym}

\usepackage{url}
\usepackage{xcolor}
\definecolor{newcolor}{rgb}{.8,.349,.1}

\usepackage{times}
\usepackage{epsfig}
\usepackage{graphicx}
\usepackage{amsmath}
\usepackage{amssymb}
\usepackage{multirow}
\usepackage{subfig}
\usepackage[export]{adjustbox}
\usepackage{enumitem}
\usepackage{algorithm}
\usepackage{algpseudocode}

\newcommand{\oc}[1]{\textcolor{black}{#1}}

\newcommand{\ie}{\textit{i.e.}}

\journal{Computer Vision and Image Understanding}

\begin{document}

\ifpreprint
\setcounter{page}{1}
\else
\setcounter{page}{1}
\fi

\begin{frontmatter}
		
\title{Learning Test-time Augmentation for Content-based Image Retrieval}

\author[1]{Osman \snm{Tursun}\corref{cor1}} 
\cortext[cor1]{Corresponding author}
\ead{osman.tursun@qut.edu.au}
\author[1]{Simon \snm{Denman}}
\author[1]{Sridha \snm{Sridharan}}
\author[1]{Clinton \snm{Fookes}}

\address[1]{Signal Processing, Artificial Intelligence and Vision Technologies (SAIVT), Queensland University of Technology, Australia}

\received{}
\finalform{}
\accepted{}
\availableonline{}
\communicated{}

\begin{abstract}
	Off-the-shelf convolutional neural network features achieve outstanding results in many image retrieval tasks. However, their invariance to target data is pre-defined by the network architecture and training data. Existing image retrieval approaches require fine-tuning or modification of pre-trained networks to adapt to variations unique to the target data. In contrast, our method enhances the invariance of off-the-shelf features by aggregating features extracted from images augmented at test-time, with augmentations guided by a policy learned through reinforcement learning. The learned policy assigns different magnitudes and weights to the selected transformations, which are selected from a list of image transformations. Policies are evaluated using a metric learning protocol to learn the optimal policy. The model converges quickly and the cost of each policy iteration is minimal as we propose an off-line caching technique to greatly reduce the computational cost of extracting features from augmented images. Experimental results on large trademark retrieval (METU trademark dataset) and landmark retrieval (ROxford5k and RParis6k scene datasets) tasks show that the learned ensemble of transformations is highly effective for improving performance, and is practical, and transferable.
\end{abstract}


\end{frontmatter}

\section{Introduction}
The internal activations of convolutional neural networks (CNNs) pre-trained on ImageNet have demonstrated astounding results when utilised for various object (image) retrieval tasks \cite{sharif2014cnn,tursun2019componet}. Such activations are termed off-the-shelf CNN features and compared to conventional hand-crafted features, they are more discriminative, compact and accessible. To this end, off-the-shelf CNN features are widely used for object retrieval over or in combination with conventional hand-crafted features.

However, un-altered off-the-shelf features are often not sufficiently robust to adapt to variations in the target data including changes in scale, illumination, orientation, color, contrast, deformations, and background clutter  \cite{aker2017analyzing,gong2014multi}. Thus image retrieval using such features can fail when these challenges are present in test data, as un-altered off-the-shelf features have no in-built invariances beyond translational invariance. This is due to these networks being predominately trained with natural images and light data augmentation. For example, ResNet \cite{He2015} was trained with only the simple data augmentations of random-crops and horizontal flips. Therefore, off-the-shelf features are not scale, rotation or contrast invariant. However, transform invariant features are desirable for challenging object retrieval tasks.

Many methods have been proposed to enhance the transform invariance of CNN features for classification and recognition. Those methods incorporate transform invariance either via special network structures, data augmentation, or both. Feature extraction modules (\ie, spatial transformer \cite{jaderberg2015spatial}, transform capsules \cite{hinton2011transforming}, multi-column networks \cite{ciregan2012multi}) and alternate filter formulations (\ie, deformable filters \cite{dai2017deformable}, transformed filters \cite{marcos2016learning,follmann2018rotationally} and pooling \cite{gong2014multi,marcos2016learning,tolias2015particular}) are types of neural structure that can improve the transform invariance of neural networks. Generally, these models require training, or are only valid for certain datasets or transformations (\ie, scale or rotation), and so lack generality. By comparison, data augmentation is a simple but effective way to achieve transform invariance as only extra transformations on input images are necessary during training or inference. This allows the generation of a more robust descriptor from multiple transformed samples, and the cost of data augmentation can be minimised by parallelization.

In this work, we propose an automatic test-time augmentation (auto TTA) that uses an ensemble of learned test-time augmentations (TTA) applied to a single input image to increase the invariance of off-the-shelf features, without sacrificing their compactness and discriminability. Our key motivation is to improve the performance off-the-shelf-features for image retrieval without network fine-tuning. Approaches that fine-tune off-the-shelf features are computationally expensive, noise-sensitive and face the risk of over-fitting and losing generality. Moreover, fine-tuning is only effective for the data on which the model is fine-tuned, and does not increase the generality of the underlying model. Existing fine-tuning-free studies for image retrieval are related to pooling \cite{gong2014multi,marcos2016learning,tolias2015particular}, feature selection/aggregation \cite{gong2014multi} or hand-crafted TTA (\ie, scaling, rotating). Compared to hand-crafted TTA, our approach exploits other data augmentations besides scaling and rotating, and requires no expert knowledge of the target image retrieval task. Additionally, the proposed auto TTA method not only learns transformations and their magnitudes, but also assigns weights to each transformation. In comparison, traditional hand-crafted TTA approaches such as multi-scale resizing only utilise a single type of transformation with pre-selected magnitudes and uniform weights.

We propose a simple but effective and efficient procedure for automating the learning process of the optimal TTA for image retrieval. Finding the optimal ensemble of TTA is a discrete search problem. Random, grid and heuristic searches are possible solutions but are expensive and impractical. Reinforcement learning-based searching has been widely used for discrete search problems \cite{zoph2016neural,cubuk2019autoaugment} such as neural architecture search (NAS) \cite{zoph2016neural}. We therefore apply a reinforcement learning-based search to find the best ensemble of TTA to extract invariant features for image retrieval. Usually, neural hyper-parameter search and NAS with reinforcement learning require enormous computational resources and are time-consuming. For example, learning data augmentations with NAS requires at least 5,000/15,000 GPU hours on the CIFAR10/ImageNet datasets as reported by \cite{cubuk2019autoaugment}. We reduce this time complexity to less than 4 GPU hours with the following improvements:
\begin{itemize}
	\item Reducing the time cost of each learning iteration. We avoid the repeated deep feature extraction process by reusing transformed features through off-line feature caching.
	\item Improving the rate of convergence during training. To this end, we proposed a metric-learning (triplet loss) based framework where a list of image transformations with customized settings and transformations for image retrieval tasks are utilised.
\end{itemize}
With this framework, the training starts to converge after approximately 2000 steps.

Our experimental results on trademark (METU trademark \cite{metudeeptursun} dataset) and landmark retrieval (ROxford5k \cite{philbin2007object,radenovic2018revisiting} and RParis6k \cite{philbin2008lost,radenovic2018revisiting} datasets) tasks show the learned ensemble of image transformations increases the performance of off-the-shelf features. The learned transformation demonstrates its transferability to off-the-shelf features from different pre-trained networks, and is applicable to various aggregation methods. We achieve SOTA comparable MAP results on the challenging METU trademark dataset with the learned ensemble of TTA.

In summary, our contribution in this paper is threefold: 

\begin{itemize}
	\item We propose a new approach, learned TTA, for boosting off-the-shelf deep features performance on content-based image retrieval (CBIR) task by learning a weighted combination of augmentations.
	\item To make this approach efficient and effective, we introduce metric-learning based training framework, off-line feature caching and a list of customised image transformations.
	\item Experimental results on trademark retrieval and landmark retrieval tasks show that the learned TTA is highly effective for improving performance, and is practical and transferable.
\end{itemize}

\section{Related Literature}
In this section, we introduce recent studies on test time augmentation (TTA) and content-based image retrieval (CBIR) with deep features.

\subsection{Test Time Augmentation}
Test-time augmentation has been widely used for image recognition and retrieval tasks \cite{gong2014multi,krizhevsky2012imagenet,simonyan2014very,szegedy2015going} due to its effectiveness. Multi-cropping and horizontal-flipping are applied when using AlexNet \cite{krizhevsky2012imagenet}, VGG16 \cite{simonyan2014very}, ResNet \cite{He2015}, and Inception \cite{szegedy2015going} models. Multi-scale cropping is introduced by \cite{gong2014multi}, and brings improvements for both image recognition and retrieval tasks. Scaling (high resolution inputs) improves performance in image retrieval tasks \cite{babenko2015aggregating,radenovic2018fine} where results with inputs at a resolution of 1,024 square are better than results with images at lower resolutions such as 512 or 256 square. Other types of data augmentation techniques such as rotation \cite{perez2018data,matsunaga2017image,wang2018test}, translation \cite{perez2018data,matsunaga2017image,wang2018test}, color shifting \cite{perez2018data,nalepa2019training} and random noise \cite{wang2018test} are also used at test-time for recognition and semantic segmentation tasks. To the best of our knowledge, all these test-time data augmentation approaches are introduced in a heuristic manner. The proposed method, by contrast, automatically generates an ensemble of TTA. It, therefore is easily utilised for other less-studied image retrieval tasks such as logo retrieval, and is useful for discovering other robust and helpful data augmentations.

In this work, we seek to learn an optimal test-time data augmentation policy through reinforcement learning. Similar approaches have been proposed for image recognition tasks to learn the optimal train-time data augmentation, but not test-time augmentation. \cite{cubuk2019autoaugment} applied a reinforcement learning method to find the best data augmentation policy. However, their method is computationally infeasible for users with limited computational resources, as the hyper-parameter search converges only after repeatedly training a deep CNN network over 15,000 times. Alternative efficient approaches reduce the computational cost by minimising the number of training runs of the large deep learning model through the training of smaller models. For example, \cite{ho2019population} applied a Population Based Training scheme to learn a candidate schedule by training a population of $k$ (\ie, 16) tiny models, then trained the target full size model with the candidate data augmentation schedule. \cite{lim2019fast} proposed a more efficient search strategy by avoiding repeated training of child models and the target model.  In contrast, we apply the same reinforcement learning based search method proposed by \cite{cubuk2019autoaugment}, and find learning a TTA scheme is computationally feasible compared to learning a set of data augmentations at training time, as no backpropagation of the deep CNN model is required to learn the ensemble of TTA. We further minimise the cost by avoiding repeated forward propagation through caching. Our proposed method, therefore, is computationally efficient, yet it still able to train for in excess of 10,000 iterations in a short period of time.

\subsection{Content-based Image Retrieval with Deep Features}
\oc{Deep CNN features are also applied to CBIR tasks following their success in other computer vision tasks. In CBIR, deep CNN features show competitive performance compared to the hand-crafted features \cite{tolias2015particular,kalantidis2016cross,tursun2019componet}. Unlike hand-crafted features, deep features are free from heavy feature engineering and have a smaller feature size. The performance of deep CNN features in CBIR tasks is further improved by two types of approaches: fine-tuning CNNs, and applying additional processes during the feature extraction step. As the proposed approach belongs to later category, here, we mainly discuss approaches which improve the performance of deep features by applying pre/post processing during feature extraction.} 

\oc{Pre-processing approaches are applied to the input image before feature extraction, while post-processing approaches are applied to the extracted off-the-shelf features. Considering post-processing approaches, special pooling and attention can be applied to the extracted deep CNN features to enhance their robustness while maintaining a compact feature size. Pooling based approaches apply spatial pooling to each channel of extracted features. For examples, MAC \cite{tolias2015particular} applies max pooling, while SPoC \cite{babenko2015aggregating} applies average pooling. GeM \cite{radenovic2018deep} is a generalised version of average and max pooling. R-MAC applies MAC pooling on multiple regions at various scales. Attention-based approaches apply spatial-wise and channel-wise weightings to the extracted CNN features prior to pooling. CRoW \cite{kalantidis2016cross} applies spatial-wise and channel-wise weights, which are calculated from the extracted deep features. \cite{Jimenez_2017_BMVC} utilises class activation maps (CAM) extracted from pre-trained networks as spatial weights. \cite{tursun2019componet} also utilises CAM but uses a fine-tuned CNN trained to identify essential components in images. The proposed method is a pre-processing approach that applies multiple data augmentations to the input image before feature extraction. Other pre-processing based approaches also apply data augmentations, such as cropping \cite{tursun2019componet} and scaling \cite{tursun2021learning,seddati2017towards}, however such approaches are heuristically defined while the data augmentations applied in our approach are learned from the data.}

\section{Proposed Method} %
In the following section, we first illustrate how to generate augmented features with a learned ensemble of test-time augmentations (TTA), and then describe how an optimal ensemble of TTA is learned.

\renewcommand{\algorithmicrequire}{\textbf{Input:}}
\renewcommand{\algorithmicensure}{\textbf{Output:}}
\begin{algorithm}[!th]
	\caption{\oc{DEEP Feature Extraction with TTA}}
	\label{algo:algo1}
	\begin{algorithmic}[1]
		\Require Image $I$, Model $f$, Ensemble of transformations $T$, Weights of transformations $W$
		\Ensure Feature $\mathbf{\hat{f}}(I)$
		\State $F^{(T)} \gets []$		
		\For{$t_j \in T$}
		\State $I' = t_j(I)$ \Comment{Apply transformation $t_j$ on $I$}
		\State $f_i = f(I', i)$ \Comment{Extract features from $i$th layer of $f$}
		\State $\hat{f}_i = post\_processing(f_i)$
		\State $F^{(T)}.append(\hat{f}_i)$
		\EndFor
		
		\State $\mathbf{\hat{f}}(I) =  weighted\_average(F^{(T)}, W)$
		\State \Return $\mathbf{\hat{f}}(I)$
	\end{algorithmic}
\end{algorithm}

\begin{figure*}[!t]
	\centering
	\includegraphics[width=0.95\linewidth]{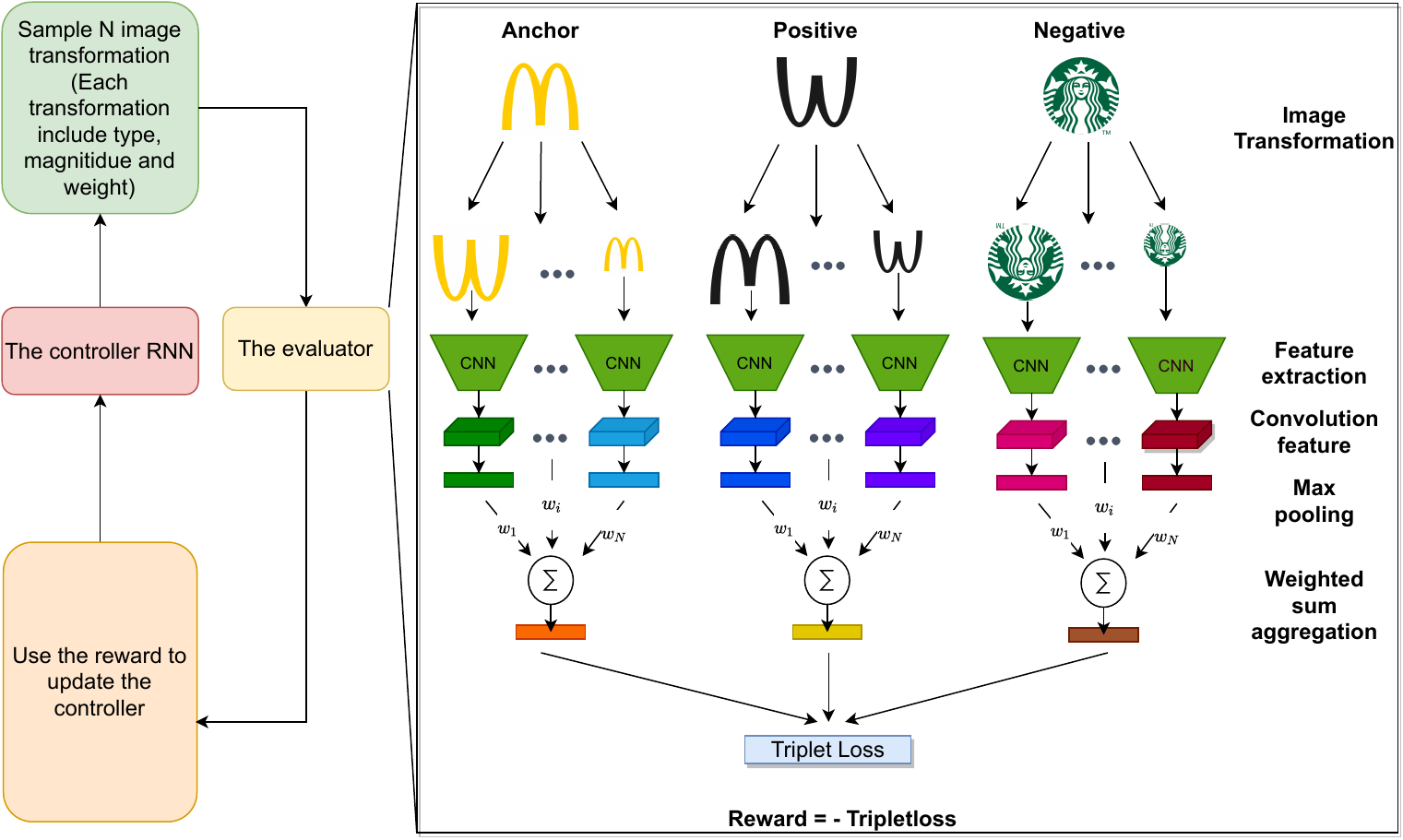}
	\caption{The proposed scheme for searching for the optimal set of test-time data augmentations to enhance off-the-shelf CNN feature robustness for image retrieval. A controller, an RNN (for details see Fig. \ref{fig:RNN}), is used to search for an ensemble of image transformations which is evaluated with a triplet network (details are given in Sec. \ref{sec:tri}). The ensemble of image transformations is applied to augment images that are sent to the triplet network for feature extraction. Extracted features are aggregated (sum aggregation), and the aggregated features is used to calculate the triplet loss whose negative value is the reward. Finally, the reward is used to update the controller.}
	\label{fig:scheme}
\end{figure*}

\begin{figure*}[!th]
	\centering
	\includegraphics[width=0.95\linewidth]{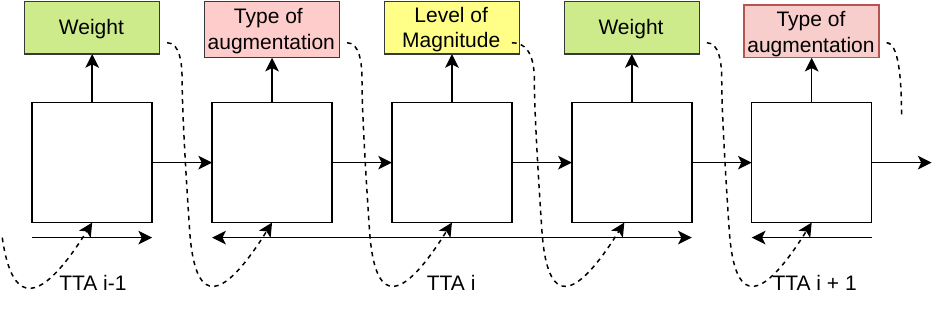}
	\caption{\oc{The structure of the controller RNN. It is a one-layer LSTM with 256 hidden nodes and which outputs $3 \times N$ softmax predictions. It generates $N$ TTA, and each of which is parameterised by a type, magnitude, and weight.}}
	\label{fig:RNN}
\end{figure*}

\subsection{Building a Robust Feature from an Ensemble of TTA}

An augmented feature $\mathbf{\hat{f}}(I)$ for image $I$ is generated by aggregating features extracted from transformed variants of $I$. \oc{This subsection descibes the feature extraction process, and the procedure is also described in Algo. \ref{algo:algo1}.}

The features used for aggregation are extracted using the following feature extraction protocol used in the recent studies of \cite{babenko2015aggregating,radenovic2018fine,kalantidis2016cross}.  Consider $f_i(I)$ to be the extracted features from the $i$th layer of a CNN when $I$ is the input image. The $i$th layer usually a convolutional or fully-connected layer. If the $i$th layer is a convolutional layer,  $f_i(I)$ is a 3D tensor of width $W$, height $H$, and channels $C$. The extracted feature is transformed into a 1D vector by reshaping, encoding or pooling. Conventionally, a pooling process such as channel-wise max pooling (MAC) \cite{azizpour2015generic} or sum pooling (SPoC) \cite{babenko2015aggregating} is applied to $f_i(I)$ to obtain a compact feature. Further, post-processing operations such as PCA whitening and normalisation may also be applied. In this work, aggregation, PCA whitening and L2 normalisation are applied to the feature $f_i(I)$, to obtain $\hat{f_i}(I)$.

Let $t_j$ represent the $j$th type of image transformation, and $T= \{t_j | 1\leq j \leq n\}$ is an ensemble of $n$ transformations. The ensemble of features $F^{(T)}$ extracted from image $I$ by applying the transforms $T$ is given by,
\begin{equation}
	\centering
	F^{(T)} = \{\hat f_i(t_j(I)) | 1\leq j \leq n\}.
\end{equation}
In this work, the final augmented feature $\mathbf{\hat{f}}(I)$ is generated through a weighted sum aggregation of $F^{(T)}$,
\begin{equation}
	\centering
	\mathbf{\hat{f}}(I) =\sum F^{(T)}= \sum_{j=1}^{n} w_i \hat f_i(t_j(I)).
\end{equation}

Finally, $\mathbf{\hat{f}}(I)$ is L2 normalised, and can be compared to other features using Euclidean distance.

In this work, the feature extraction during training and inference follows the same aforementioned protocol. In our experiments, the MAC pooling of the last activation layer of the Conv5 block of VGG16 \cite{simonyan2014very} is used.  

\subsection{Learning an Ensemble of Image Transformations}

In this work, finding the best ensemble of image transformations for off-the-shelf feature augmentation is formulated as a discrete search problem. Inspired from the work by \cite{cubuk2019autoaugment}, we apply Reinforcement Learning (RL) to determine the optimal ensemble of TTA for the target image retrieval task. This approach finds the best policy, $S$, which includes $n$ learned image transformations, and comprises the transformation operations and their corresponding magnitudes and weights. The range of magnitudes is discretised into 10 values. Weights are also discretised into 10 levels: 0 to 10, but their L1 normalised values are applied during weighted sum aggregation. To this end, if we sample $N$ image transformations, the search space would be $(16 \times 10 \times 10)^n$. In this work, $n$ is set to 8, as setting $n$ too large increases the computational cost, while using too small a value of $n$ decreases the diversity of the learned transformations which has a negative impact on performance.

\oc{A recurrent neural network (RNN) based controller is trained to sample policies, as each policy is a sequence of augmentations. The controller is a one-layer LSTM \cite{hochreiter1997long} with 256 hidden units and $3 \times N$ softmax predictions for the $N$ predicted transformations and their corresponding magnitudes and weights.}
At each step, the RNN produces a decision using a softmax classifier and sends it to the next step as an input for the next decision. In total, the controller has $3 \times N$ softmax predictions. The final sequence of decisions is the policy $S$, which is sent to the evaluator for evaluation. The evaluation score is used as the reward, $R$. The controller is updated with $R$. \oc{An overview of this process is given in Fig. \ref{fig:scheme}, and the structure of controller is described in Fig. \ref{fig:RNN}.}

\begin{table*}[]
	\begin{minipage}{\textwidth}
		\setcounter{mpfootnote}{1}
		\caption{List of all image transformations used in this work. The majority of the data augmentations are the same data augmentations introduced by \cite{cubuk2019autoaugment}, though new operations and new settings are introduced for CBIR task. Operations shown with bold font are newly added data augmentations, while underlined operations are contain different settings. Note all these transformations are included in the PIL$^{a}$ library.}
		\begin{center}
				\begin{tabular}{l | l | p{12cm} | p{2cm} }\hline
					\bf id & \bf Operation & \bf Name Description & \bf Range of Magnitudes \\\hline
					1 & \bf{Resize}    & Resize the largest side of the image to the magnitude size with preserving the original aspect ratio. Two different ranges are applied for trademark and landmark images.                           & [64, 352] (trademarks),  [384, 1536] (landmarks)  \\\hline
					2 & \underline{Rotate} & Rotate the image by the rate magnitude degrees. & [-180, 162] \\\hline
					3-4 & ShearX (Y) & Shear the image along the horizontal (vertical) axis by the rate magnitude. & [-0.3, 0.3] \\\hline
					5-6 & \underline{TranslateX (Y)} & Translate the image in the horizontal (vertical) direction by	the multiplication of the magnitude with the width (height). The input image is extended by reflecting the edge of last pixels. & [-0.45, 0.36] \\\hline
					7 & AutoContrast & Maximise the image contrast, by making the darkest pixel black and lightest pixel white. & \\\hline
					8 & Invert & Invert the pixels of the image.  & \\\hline
					9 & Equalise & Equalise the image histogram. & \\\hline
					10 & Solarise & Invert all pixels above a threshold value of magnitude. & [0, 256] \\\hline
					11 & \underline{Posterise} & Reduce the number of bits for each pixel to magnitude bits. & [1, 8] \\\hline
					12 & Contrast & Control the contrast of the image. A magnitude=0 gives a gray image, whereas magnitude=1 gives the original image. & [0.1, 1.9] \\\hline
					13 & Colour & Adjust the colour balance of the image, in a manner similar to the controls on a colour TV set. A magnitude=0 gives a black white image, whereas magnitude=1 gives the original image. & [0.1, 1.9] \\\hline
					14 & \underline{Brightness} & Adjust the brightness of the image. A magnitude=0 gives a black image, whereas magnitude=1 gives the original image. & [0.4, 1.9] \\\hline
					15 & Sharpness & Adjust the sharpness of the image. A magnitude=0 gives a blurred image, whereas magnitude=1 gives the original image. & [0.1, 1.9] \\\hline
					16 & \textbf{Horizontal-flip} & Flip the image in horizontal direction.                       & \\\hline
					17 & \textbf{Contour} & Extract the contour of the image.                       & \\
					\hline
				\end{tabular}
			\end{center}
			\label{tab:imgtrans}
			\footnotesize{$^{a}$ python-pillow.org}
		\end{minipage}
	\end{table*}
	
	\subsubsection{Evaluating the Learned TTA Policy}
	\label{sec:tri}
	The TTA policy generated by a controller is evaluated with the evaluator module displayed in the Fig. \ref{fig:scheme}. The evaluator assigns a high reward for a valid TTA policy and vice versa. A valid TTA policy should be capable of generating distinctive augmented features from off-the-shelf features extracted using a pre-trained neural network. \oc{Distinctive features for image retrieval will keep similar images close and dissimilar images distant in the feature space. To this end, we applied a triplet network \cite {schroff2015facenet}, which is widely used for deep metric learning. As shown in the evaluator module in Fig.} \ref{fig:scheme}, the evaluator is a triplet network with three branches: anchor, positive and negative. \oc{Features extracted from the anchor and positive branches are should be similar, while features extracted from the anchor and negative branches should be dissimilar.} These three branches will share weights and are not updated during training. The same learned transformations are applied to the anchor, positive and negative images of the triplet before sending them to the corresponding branches for feature extraction. In this work, VGG16 is the backbone network of the branches. The final augmented feature is a weighted sum of the $N$ features from $N$ transformed images. Augmented features are extracted for all triplets from the training dataset with a Triplet network. The final reward, $R$, is given by,
	
	\begin{equation}
		R = -\frac{1}{k}\sum_{i=1}^{k} max(||\mathbf{\hat f}(I_i^a) - \mathbf{\hat f}(I_i^+)||_2 - ||\mathbf{\hat f}(I_i^a) - \mathbf{\hat f}(I_i^-)||_2 + \alpha, 0),
	\end{equation}
	where $k$ is the total number of triplets, and $I_i^a, I_i^+, I_i^-$ is the $i$th triplet. $\alpha$ is the margin, and it is set to 0.2 as \cite{schroff2015facenet} recommended.

	\subsubsection{The Collection of Image Transformations} We create a collection of image transforms that includes the following 17 transformations: \textit{Resize, Rotate, ShearX/Y, TranslateX/Y, AutoContrast, Invert, Equalise, Solarise, Posterise, Contrast, Colour, Brightness, Sharpness, Horizontal-flip and Contour}. It includes most of the image transformations collected by \cite{cubuk2019autoaugment} except for Cutout and Sample pairing, which are not useful for CBIR tasks. In addition, we add three new promising transformations: Resize, Horizontal-flip and Contour. These three additional transformations have been heuristically used as test-time augmentations in prior works. For example, Horizontal-flip is used for augmenting the test set in image classification tasks \cite{krizhevsky2012imagenet,simonyan2014very} while Resize is used for generating multi-scale features in landmark image retrieval \cite{radenovic2018revisiting} and trademark retrieval \cite{tursun2021learning} tasks. We introduce the Contour (or edge) operation as contour based shape matching is robust to illumination changes and domain difference in image retrieval tasks \cite{radenovic2018deep}.

	All these data augmentations are implemented with the popular Python image library, PIL \footnote{https://pillow.readthedocs.io/}. We use the same operation magnitude ranges suggested by \cite{cubuk2019autoaugment}, except for Rotate, Translate X/Y, Posterise and Brightness. \oc{We increased the range of the Rotate operation, so it will cover upside down cases. We updated the ranges of Translate X/Y, Posterise and Brightness to prevent images appearing similar after the augmentation step. For example, images become black when the Brightness is set to zero.}  A detailed description of these transformations and their magnitude ranges are given in Table \ref{tab:imgtrans}.

	\subsubsection{Accelerated Training via Off-line Caching}
	Compared to other works \ie, neural architecture search (NAS) \cite{cubuk2019autoaugment,zoph2018learning}, one iteration of the proposed method takes a comparatively small amount of time as the feature extraction network is fixed (i.e. it is not updated during training). However, to further speed up the process, we cache features from previous iterations such that features are extracted only once for each data transformation. We load features for each data augmentation from the hard disk in a sequential order and process aggregation operations in-place, so  memory cost does not increase with the number of transformations applied. This simple approach accelerates the whole training process by a factor of 100. With a GeForce GTX 1080 graphic card, a single training iteration takes around $3-5$ seconds on our training dataset.
	
	\subsubsection{Training the Controller} 
	We optimise the controller with the Proximal Policy Optimisation algorithm \cite{schulman2017proximal}, inspired by \cite{cubuk2019autoaugment,zoph2018learning}. The controller is trained with a learning rate of $10^{-4}$. In total, the controller samples about 10,000 policies (the mode converges before 10,000 steps). The policy generated by the converged model is selected for inference.
	
\section{Experiment}
\label{sec:exp}
We tested the proposed method on both trademark and landmark image retrieval tasks, which are challenging and well-known content-based image retrieval tasks. In the following sub-sections, we present the datasets, evaluation metrics, experimental setup, and results.

\subsection{Datasets}  

\begin{figure}[!t]
	\centering
	
	\includegraphics[width=\linewidth]{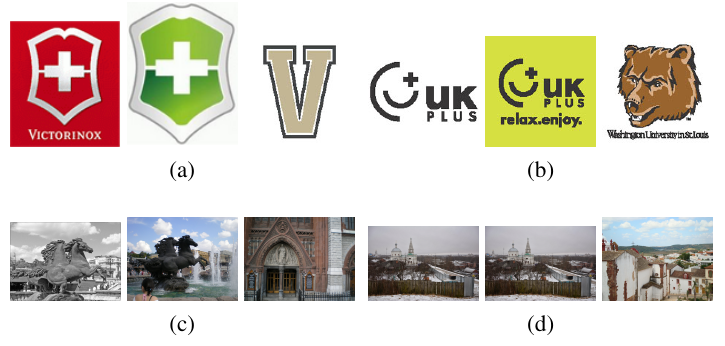}
	
	\caption{Examples of triplets from the METU trademark and the Google landmark datasets. (a-b) are trademark triplets, while (c-d) are landmark triplets. Individual images in each group are, from left to right: anchor, positive and negative.} 
	\label{fig:pairs}
\end{figure}

In this study, The proposed method is evaluated with two well-known CBIR tasks: trademark retrieval and landmark retrieval. Two different datasets are selected from the same domain for training and testing for each task. For example, the NPU trademark dataset \cite{lan2017similar} is selected as the training dataset, and the METU dataset \cite{metudeeptursun} is selected as the testing dataset for trademark image retrieval. NPU is a small dataset that includes 317 similar trademark groups and each group contains at least two similar trademarks. In total, 100,000 trademark triplets are created with these similar groups. In each triplet, the anchor and positive image are drawn from one of the 317 similar groups to obtain a similar image pair, while the negative image is randomly selected from a different group (i.e. one of the other 316 groups). Example triplets are shown in (a-b) of Fig. \ref{fig:pairs}. In comparison, the METU dataset includes nearly one million trademarks. We also created a small subset of the METU dataset by selecting 10K random images from it. This subset is used for ablation studies.

Similarly, a small subset of Google Landmark Dataset \cite{noh2017large} is used for training, and the ROxford5k \cite{philbin2007object,radenovic2018revisiting} and the RParis6k Datasets are used  \cite{philbin2008lost,radenovic2018revisiting} are selected for testing in the landmark retrieval task. The revised Oxford5k dataset includes 4,993 images (includes 70 queries), and the Paris6k dataset has 6,322 images (includes 70 queries). Both datasets have three evaluation setups of different difficulty (easy, medium and hard).

We also build 100,000 landmark triplets for training. To do this, we first created 2,442 similar landmark pairs through grouping two randomly selected images from each of the most frequently appearing 2,442 unique landmarks in the Goolge Landmark dataset. Each landmark triplet is composed of anchor and positive images belong the same similar pair, and a negative image from a different similar pair. Example triplets are shown in (c-d) of Fig. \ref{fig:pairs}.

\subsection{Evaluation Metrics}
We adopt the evaluation metric, \textit{mean average precision} (MAP), shown in Equation \ref{eq:map}, which is widely used in related literature \cite{metudeeptursun,radenovic2018fine}. It is calculated using ranking positions obtained by sorting the image to query similarity distance in descending order. Note the MAP value is calculated using the top 100 ranking results for the METU trademark dataset, while for both Oxford5k and Paris6k datasets MAP is calculated from top 10 retrieved results.

\begin{equation}
	MAP@K = \frac{1}{Q}\sum_{i=1}^{Q}\frac{1}{E_i}\sum_{j}^{k}\frac{c_j}{j},
	\label{eq:map}
\end{equation}
where $Q$ is the number of queries, and $E_{i}$ is the number of expected results of the query $i$. $r_j$ is one when image with rank $j$ is the expected image, and otherwise is zero.

		\subsection{Experiment Setups}
		Images are resized before being passed to the networks for feature extraction. All trademark images from the METU dataset are resized $224 \times 224$, and all landmark images from Oxford5k and Paris6k datasets are resized to a maximum $1024 \times 1024$. The aspect ratio is kept the same as the original in during the resize operation.  To achieve this, padding is applied to the METU dataset, while no padding is applied for the landmark images.
		
		All experiments were conducted on a PC with a single NVIDIA 1080 Ti graphic card, a 3.40GHz Intel Core i7-6700 CPU, and 32Gb RAM.
		
		\subsection{Results and Analysis}
		
\begin{figure*}[!t]
			\centering
			\includegraphics[width=\linewidth]{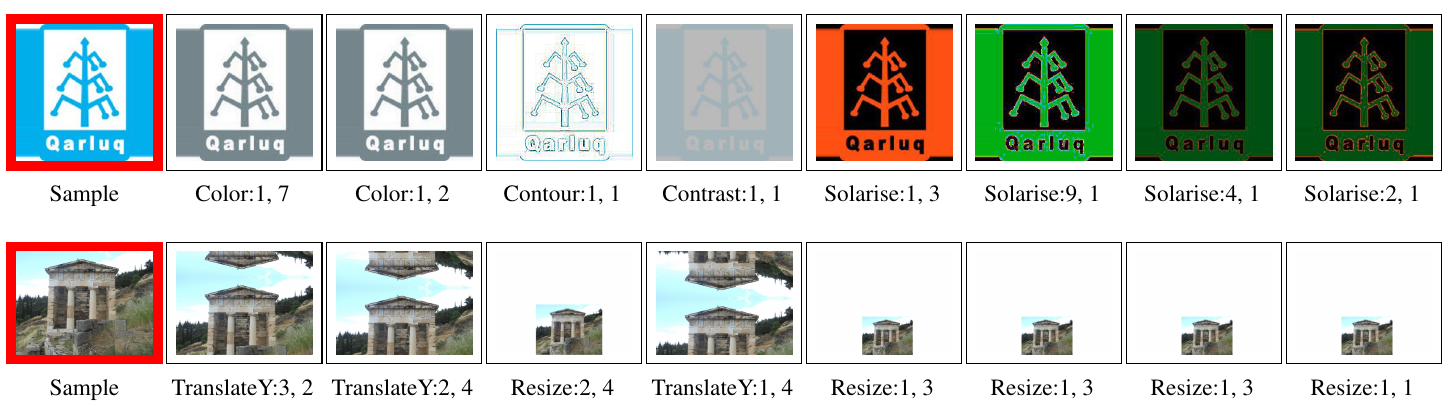}
			
			\caption{Visualisation of the learned ensemble of TTA with a sample image. Under each transformed image, the applied image transformation, magnitude and weight are given. The top row shows the ensemble for trademark retrieval, while the bottom row shows the ensemble for landmark retrieval. Note that the resize operation shows only an approximate transform in consideration of the page limit and figure readability.}
			\label{fig:policy}
		\end{figure*}

		We selected the policy sampled by the controller after the policy converged on the training datasets. The same policy is used for our main evaluation and ablation studies. The policy learned for trademarks on the NPU dataset is: \textit{Color: (1, 7), Color: (1, 2), Contour: (1, 1), Contrast: (1, 1), Solarise: (1, 3), Solarise: (9, 1), Solarise: (4, 1), Solarise: (2, 1)}, and the policy learned on the sub-set of Google landmark dataset for landmark images is: \textit{TranslateY: (3, 2), TranslateY: (2, 4), Resize: (2, 4), TranslateY: (1, 4), Resize: (1, 3), Resize: (1, 3), Resize: (1, 3), Resize: (1, 1)}.
		Visualised policies with a sample image are shown in Fig. \ref{fig:policy}. Of note, the controller has learned two different policies for two different tasks.

		\captionsetup[subfigure]{labelformat=empty}
		
		{\bfseries{Network Transferability}} We tested the learned policies with several ImageNet pre-trained networks including VGG16 \cite{simonyan2014very}, ResNet \cite{He2015}, DenseNet121 \cite{huang2017densely} and AlexNet \cite{krizhevsky2012imagenet}. All pre-trained models are from TorchVision \footnote{https://pytorch.org/docs/stable/torchvision/models.html}. The feature extraction layers and results are shown in Table \ref{tab:net}. With the learned TTA, all MAP results have increased significantly compared to those without TTA. The learned TTA, therefore, has a good network transferability.

		{\bfseries{Aggregation Transferability}} %
		Here, we tested the learned policy with other types of aggregation methods, namely SPoC \cite{babenko2015aggregating}, GeM \cite{radenovic2018fine}, CRoW \cite{kalantidis2016cross} and R-MAC \cite{tolias2015particular}. Features extracted from the last activation layer of Conv5 of VGG16 are used for implementing these aggregation methods. Hyper-parameters used in these methods are same to those in \cite{radenovic2018fine}.

		From the results shown in Tab. \ref{tab:agg}, we can see the overall MAP results of all aggregation methods are improved. This shows the learned TTA is transferable to other aggregation methods. Nevertheless, we noticed MAC pooling achieves the largest improvement compared to other aggregations. A further improvement might be achievable for other aggregations if the same type of aggregation is applied during training and testing, as is the case for MAC aggregation.

			\begin{table*}[!t]
				\caption{Results of off-the-shelf features from ImageNet pre-trained models with/without the learned ensemble of TTA. Results of learned image transformations are highlighted in bold if improved.}
				\begin{center}
						\begin{tabular}{c | c | c |c | c | c | c| c| c| c| c}
							\hline
							\multirow{2}{*}{\bf Networks} & \multirow{2}{*}{\bf Layer} & \multirow{2}{*}{\bf Dim} & \multirow{2}{*}{\bf TTA} & \bf METU-10K & \multicolumn{3}{c|}{\bf ROxford (MAP@10)} & \multicolumn{3}{c}{\bf RParis (MAP@10)} \\ \cline{6-11}
							
							&  &  &  & \bf (MAP@100) & \bf Easy & \bf Medium & \bf Hard & \bf Easy & \bf Medium & \bf Hard \\\hline

							\multirow{2}{*}{VGG16 } &  \multirow{2}{*}{Conv5} & \multirow{2}{*}{256} &  & 33.49  & 55.22 & 58.57 & 30.29 & 89.15 & 92.00 & 62.00\\\cline{4-11}
							&  &  & \checkmark & 
							\bf 41.86 & \bf 67.72& \bf 69.57& \bf 34.29 & \bf 91.57 & \bf 95.29 & \bf 73.14\\\hline
							
							\multirow{2}{*}{VGG16 } &  \multirow{2}{*}{Conv5} & \multirow{2}{*}{512} &  & 28.71  & 58.24 & 59.4 & 29.49 & 89.42 & 92.86 & 64.29\\\cline{4-11}
							&  &  & \checkmark & \bf 40.29 & \bf 69.52 & \bf 68.76 & \bf 34.71 & \bf 91.62 & \bf 95.14 & \bf 71.86\\\hline
							
							\multirow{2}{*}{AlexNet } &  \multirow{2}{*}{Conv5} & \multirow{2}{*}{256} &  &  28.69 & 44.19 &  41.9 & 13.43 & 83.43 & 85.71 & 46.14\\\cline{4-11}
							&  &  & \checkmark & \bf 38.42 & \bf 52.94 & \bf 52.14 & \bf 21.00 & \bf 91.14 & \bf 92.43 & \bf 69.00\\\hline
							
							\multirow{2}{*}{ResNet50 } &  \multirow{2}{*}{Pool5} & \multirow{2}{*}{512} &  & 42.20 &  40.25  & 43.1 &  18.43 & 80.67 & 85.29 & 51.43\\\cline{4-11}
							&  &  & \checkmark & \bf 48.01 & \bf 59.04 & \bf 59.22 & \bf 31.29 & \bf 91.71 & \bf 94.00 & \bf 74.00 \\\hline
							
							\multirow{2}{*}{ResNet50 } &  \multirow{2}{*}{Pool5} & \multirow{2}{*}{2048} &  & 39.80  & 44.43 & 48.57 & 20.29 & 86.71 & 90.00 & 51.29\\\cline{4-11}
							&  &  & \checkmark & \bf 48.76 & \bf 63.87 & \bf 63.1 & \bf 33.00 & \bf 93.14 & \bf 96.29 & \bf 73.57\\\hline
							
							\multirow{2}{*}{DenseNet121 } &  \multirow{2}{*}{DenseBlock4} & \multirow{2}{*}{512} &  & 38.43  & 38.35 & 41.14 & 11.07 & 85.71 & 87.57 & 52.86 \\\cline{4-11}
							&  &  & \checkmark & \bf 45.91 & \bf 63.04 & \bf 59.50 & \bf 23.68 & \bf 93.43 & \bf 95.43 & \bf 74.29\\\hline
							
							\multirow{2}{*}{DenseNet121 } &  \multirow{2}{*}{DenseBlock4} & \multirow{2}{*}{1024} &  & 37.31  & 50.31 & 50.12 & 18.62 & 88.14 & 90.29 & 61.14\\\cline{4-11}
							&  &  & \checkmark & \bf 45.16 & \bf 69.68 & \bf 65.48 & \bf 29.29 & \bf 94.57 & \bf 96.71 & \bf 79.29\\\hline

							\end{tabular}
						\end{center}
						\label{tab:net}
					\end{table*}

					\begin{table*}[!t]
						\caption{Results of off-the-shelf features from ImageNet pre-trained models with/without the learned ensemble of TTA. Results of learned image transformations are highlighted in bold if improved.}
						\begin{center}
								\begin{tabular}{ c |c | c | c | c| c| c| c| c}
									\hline
									\multirow{2}{*}{\bf Dim} & \multirow{2}{*}{\bf TTA} & \bf METU-10K & \multicolumn{3}{c|}{\bf ROxford (MAP@10)} & \multicolumn{3}{c}{\bf RParis (MAP@10)} \\ \cline{3-9}
									
									&  & \bf (MAP@100)& \bf Easy & \bf Medium & \bf Hard & \bf Easy & \bf Medium & \bf Hard \\\hline
									
									\multirow{2}{*}{MAC} &  & 33.49  & 55.22 & 58.57 & 30.29 & 89.15 & 92.00 & 62.00\\\cline{2-9}
									& \checkmark & \bf 41.86 & \bf 67.72& \bf 69.57& \bf 34.29 & \bf 91.57 & \bf 95.29 & \bf 73.14\\\hline
									
									\multirow{2}{*}{SPoC} &  & 32.99  & 62.23 & 61.38 & 23.36 & 90.00 & 93.00 & 65.57\\\cline{2-9}
									& \checkmark & \bf 40.13 & \bf 62.46 &\bf 65.90 & \bf 30.71 & \bf 92.43 & \bf 95.29 & \bf 67.86\\\hline
									
									\multirow{2}{*}{R-MAC} &  & 38.06  & 69.24 & 64.52 & 27.50 & 92.57 & 94.57 & 74.86\\\cline{2-9}
									& \checkmark & \bf 44.93 & 68.72 & \bf 66.22 & \bf 30.31 & \bf 92.71 & \bf 95.86 & 72.14\\\hline
									
									\multirow{2}{*}{GeM} &  & 35.29  & 63.76 & 64.29 & 30.79 & 91.29 & 94.29 & 75.71\\\cline{2-9}
									& \checkmark & \bf 43.02 & \bf 69.94 & \bf 71.43 & \bf 34.71 & \bf 91.86 & \bf 95.00 & 73.86\\\hline
									
									\multirow{2}{*}{CRoW} &  & 33.15  & 67.07 & 68.76 & 34.29 & 91.00 & 93.71 & 72.00 \\\cline{2-9}
									& \checkmark & \bf 40.92 & \bf 70.74 & \bf 69.43 & 32.60 & \bf 92.00  & \bf 94.71 & 71.43 \\\hline
									
								\end{tabular}
							\end{center}
							\label{tab:agg}
						\end{table*}

\begin{figure}[]
											\centering
											\subfloat[(a) trademark]{\includegraphics[width=\linewidth]{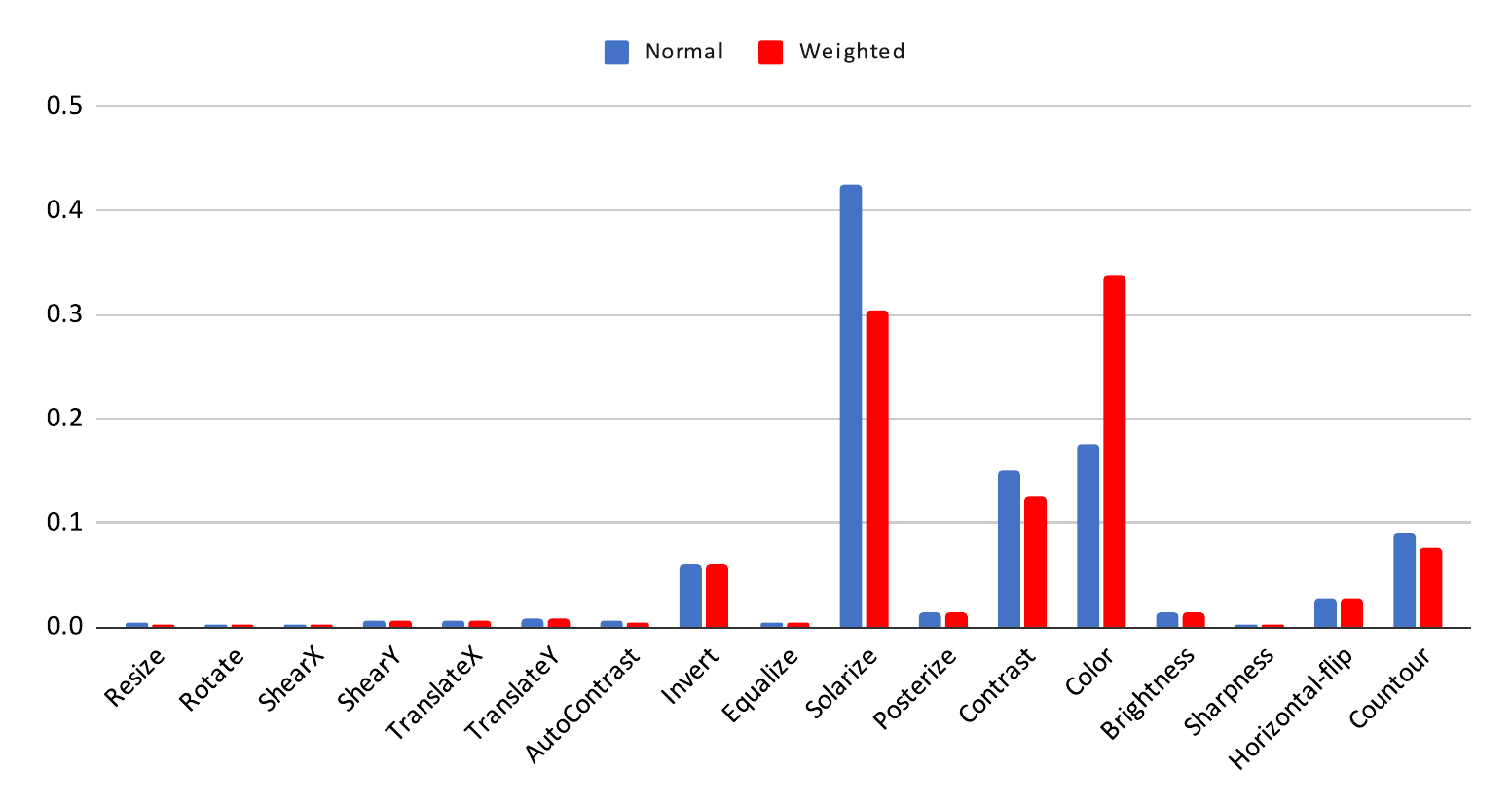}
											}
											
											\subfloat[(b) landmark]{\includegraphics[width=\linewidth]{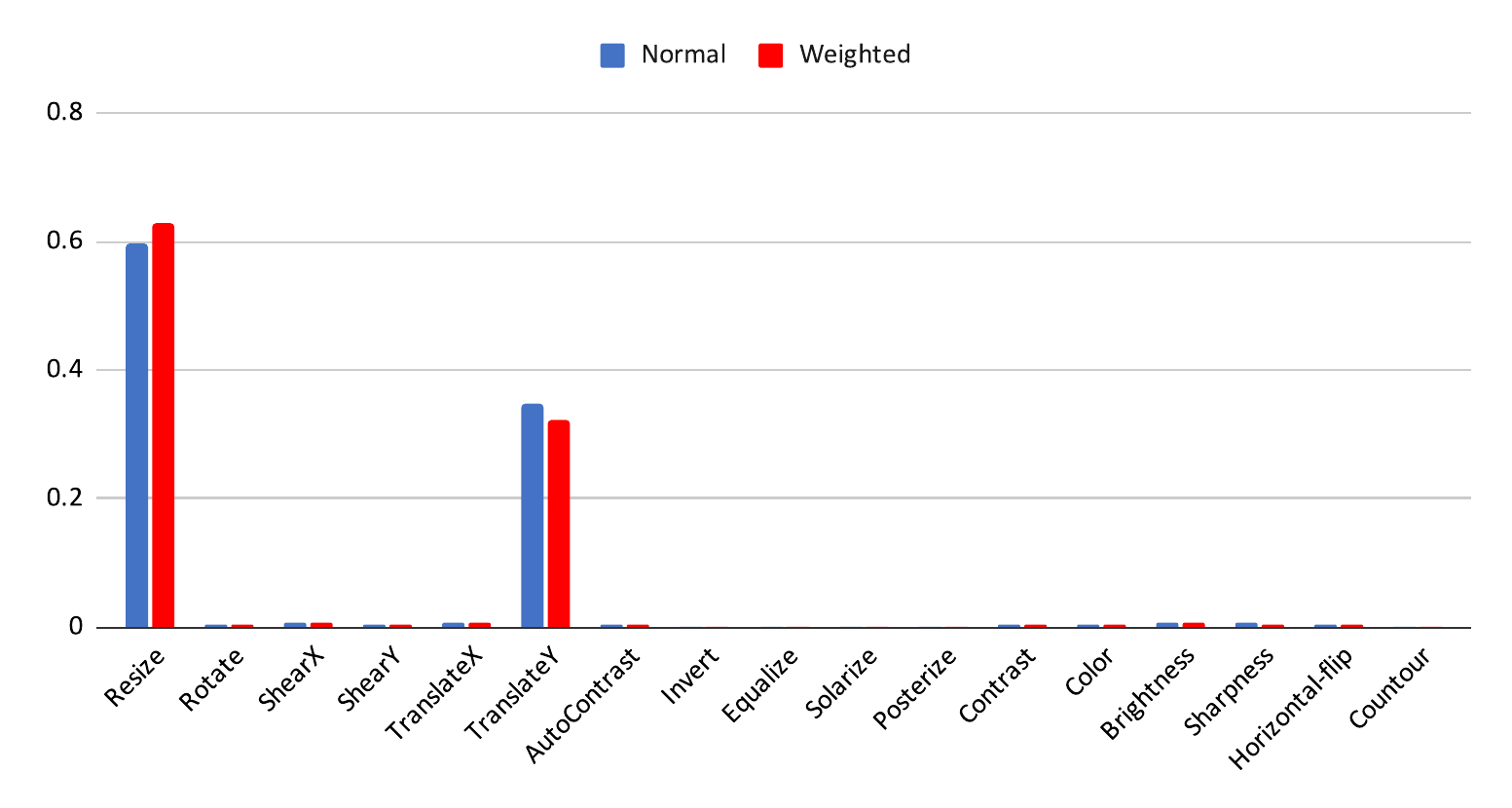}
											}
											\caption{The normal/weighted occurrence rate of image transformations in unique policies sampled during training on NPU trademark and Google landmark datasets.}
											\label{fig:toppolicy}
										\end{figure}

										{\bfseries{Useful Image Transformations and the Role of Weights}} We calculate both normal and weighted occurrence rates of all transformations in unique policies sampled during training on the trademark and landmark tasks. These results are shown in (a-b) Fig. \ref{fig:toppolicy}.  The most frequently appearing transformations also appear in the best-learned policy. For instance, in the landmark task, Resize and TranslateY operations the most frequently appearing operations. The resize operation corresponds to multi-scale operations applied in the landmark retrieval task. This finding further proves that the proposed methods learn valuable transformations. In comparison, TranslateY has not been used in any landmark retrieval studies.  The TranslateY operations in the best policy direct attention to the roof region of the buildings by removing the bottom part of the building and reflecting the upper portion. This may be a result to the roof portion of buildings having more discriminative features. On the other hand, in the trademark retrieval task, operations such as Invert, Solarise, Contrast, Color, Horizontal-flip and Contour frequently appear in the sampled policies. However, the resize and rotate operations have low occurrence rates. This is because the training data does not have scaling and rotation variance.
										
										The difference between normal and weighted frequency shows the learned weights also contribute to the performance. For example, in the landmark retrieval task, resize operations get more weight compared to the TranslateY operations. In the trademark retrieval task, the Color operations with magintude one get the largest weight. Note the Color opertions with magnitude one only turn images into grayscale. Here, we can clonclude that the learned weights ehcance the contribution factor of the most important image transformation. These finding conform to the reults of previous studies. For instance, multi-scaling have been widely used in landmark retrieval\cite{radenovic2018revisiting}, and color-based features demonstrate poor performance compared to other shape-based features in trademark retrieval \cite{tursun2015metu}. 
										
										The learned policy for trademark retrieval assigns more importance to the shape similarity of trademarks than their color similarity.
										
\begin{figure}[]
											\centering
											\includegraphics[width=\linewidth]{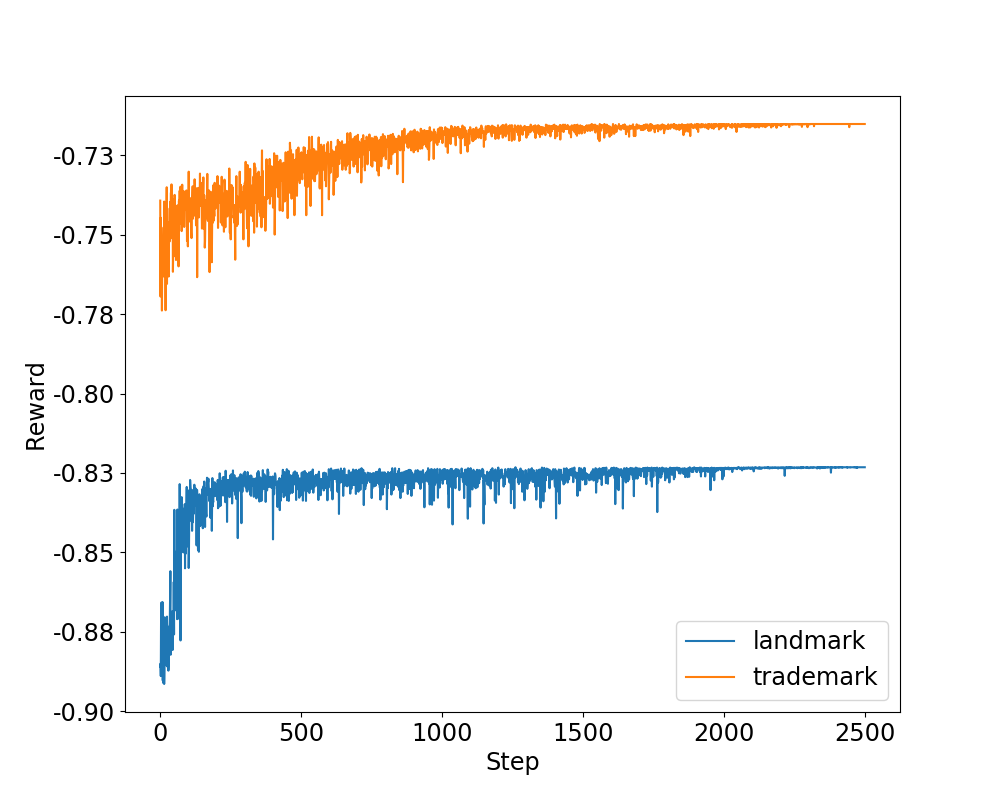}
											\caption{The reward curves during training for landmark and trademark datasets. The model starts to converge after 2,000 steps on both tasks.}
											\label{fig:reward}
										\end{figure}

										{\bfseries{Network Convergence Speed}} On both landmark and trademark retrieval tasks, the proposed method converges after 2,000 iterations as shown in Fig. \ref{fig:reward}. In our settings, each iterations takes less than 5 seconds. The training, therefore, takes three to four hours on average.
										
											{\bfseries{Feature Extraction Time}}
											The proposed method processes multiple test-time data augmentations and extracts features for each type of data augmentation. If this process is done in a sequential manner this will increase the processing time by a factor of approximately $N$, where $N$ data augmentations are applied. In this work, the maximum $N$ is 8, and the CPU processing time of our method with sequential processing is on average 0.08(s) per image,  which is around 10 times the baseline time of 0.008(s). However, we reduced the processing time to 0.02 (s) by parallelizing data augmentations and feature extraction.
											
											{\bfseries{Comparison with State-of-the-art Results}}
											We also tested the learned TTA on the full-scale METU trademark dataset. The results are show in Table \ref{tab:sota}. With the TTA applied off-the-shelf ResNet50/Pool4 MAC features, a MAP@100 score of 30.5 is achieved when the feature size is 512. The score decreases to 28.6 if the feature size is reduced to 256. These results outperform all previous SOTA results except for the recent method of \cite{tursun2021learning}. In \cite{tursun2021learning}, hand-crafted TTA (namely multi-scale resizing), regional attention and a modified R-MAC are applied. In comparison, we get a comparable results using only the learned TTA and MAC aggregation.
											
\begin{figure*}[]
														\centering
														\includegraphics[width=0.95\linewidth]{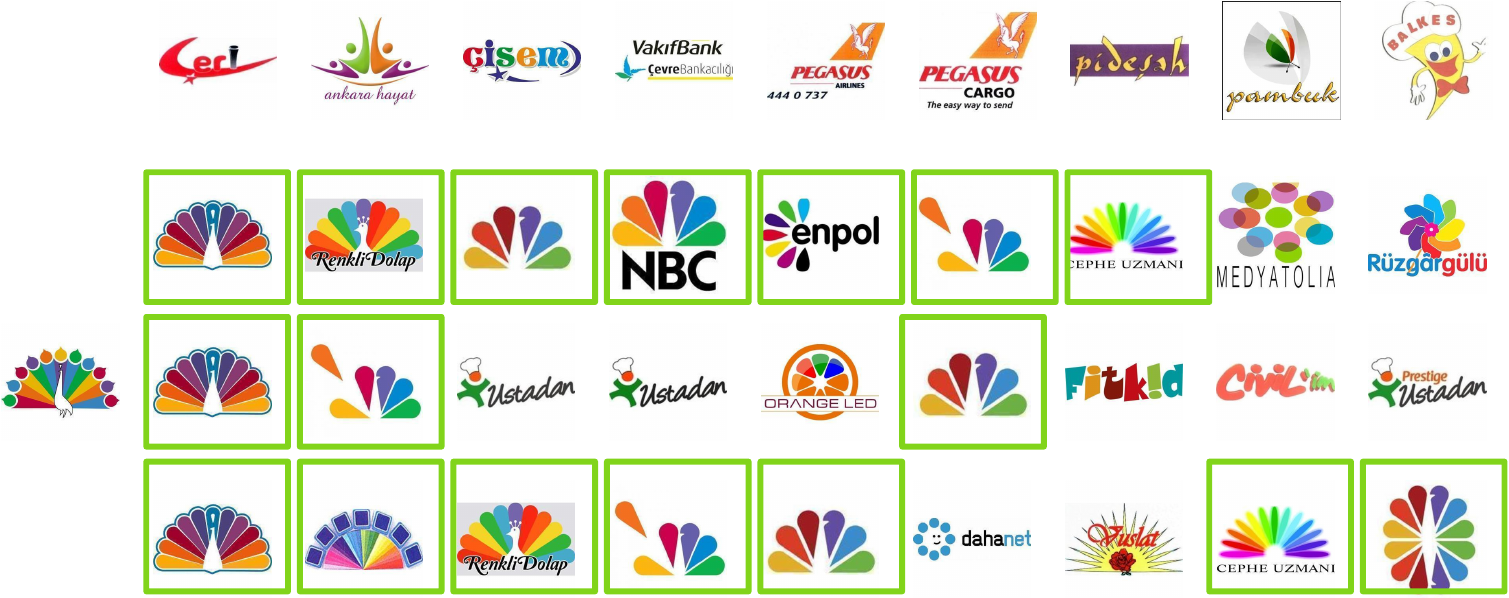}
														\includegraphics[width=0.95\linewidth]{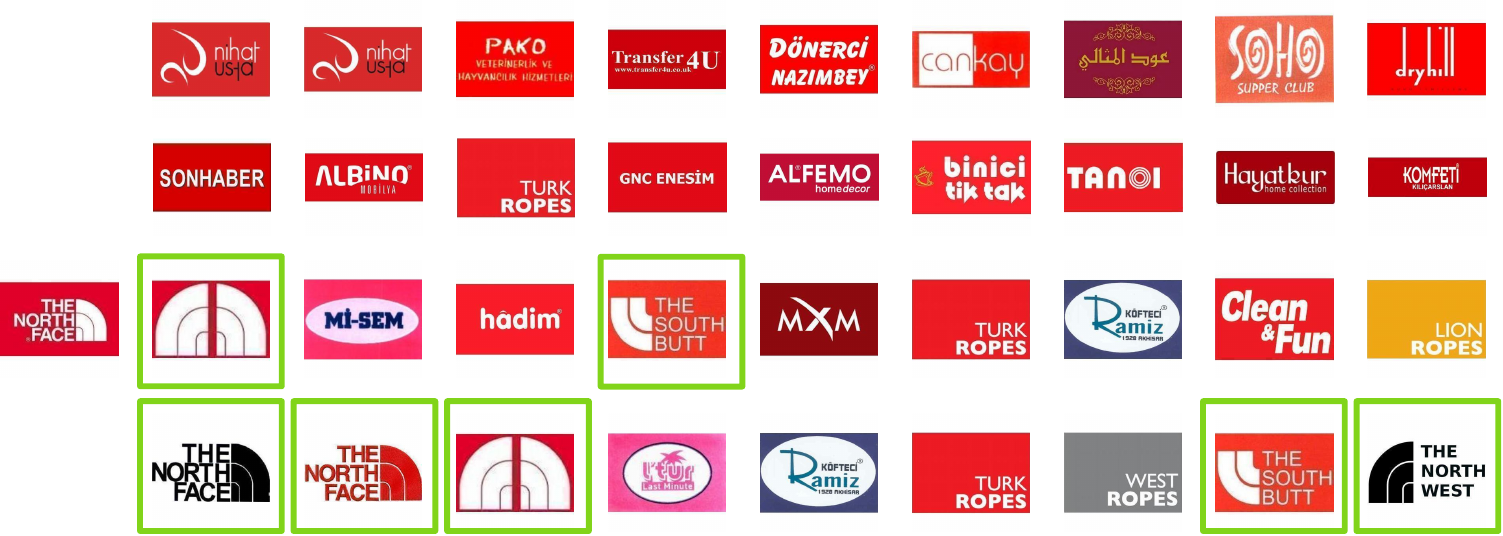}
														\caption{Comparison of top 10 results of our best method (ResNet50/Pool4 MAC with/without TTA) and the previous state-of-the-art methods (ATR MAC and MR-R-SMAC w/UAR) on the METU trademark dataset. For each query, the first row is the result of ATR MAC, the second row is MR-R-SMAC w/UAR, the third row is our method without TTA, and the last row is our method with TTA. Positive results are noted with a green box.}
														\label{fig:metuqual}
													\end{figure*}

											A qualitative comparison with the previous two SOTA methods (ATR MAC \cite{tursun2019componet} and MR-R-SMAC w/UAR \cite{tursun2021learning}) on two challenging queries is shown in Fig. \ref{fig:metuqual}. In both cases, the proposed method shows outstanding results compared to two SOTA methods and the proposed approach without TTA. Although, the MR-R-SMAC w/UAR method shows comparable results for the first query, it fails on the second query. This might be caused by the rare red background color in the METU trademark dataset. However, the proposed method is still able to return good results because of the grayscale transformations applied from the learned TTA.

												\begin{table}[h!]
													\small
													\begin{center}
															\begin{tabular}{l c  c}
																\hline
																\bf Method & \bf DIM $\downarrow$ & \bf MAP@100 $\uparrow$\\ \hline
																SPoC \cite{babenko2015aggregating} &  256 & 18.7\\ 
																CRoW \cite{kalantidis2016cross} 	& 256    & 19.8     \\
																R-MAC \cite{tolias2015particular}  		& 256 	& 24.8 \\
																MAC \cite{tolias2015particular} 	& 512  & 21.5 \\
																Jimenez \cite{Jimenez_2017_BMVC}   & 256 	& 21.0  \\
																CAM MAC \cite{tursun2019componet}  &  256   &22.3  \\
																ATR MAC \cite{tursun2019componet} &  512 & 24.9 \\
																ATR R-MAC \cite{tursun2019componet} &	256  & 25.7 \\
																ATR CAM MAC \cite{tursun2019componet}	&  512  &  25.1 \\
																MR-R-SMAC w/UAR \cite{tursun2021learning} &  256  & \textbf{31.0} \\\hline
																Ours &  512 & 30.5 \\
																(ResNet50/POOL4 MAC)						  &  256 & 28.6 \\\hline
															\end{tabular}
														\end{center}
														\caption{Comparison with the previous state-of-the-art results on the METU dataset. NAR is the normalised average rank metric.}
														\label{tab:sota}
													\end{table}

\section{Conclusion}
In this work, we applied a reinforcement learning approach to learn an ensemble of test-time data augmentations to compose robust features for image retrieval in a computationally efficient and affordable manner. The proposed method shows improvements in the performance of off-the-shelf features, and demonstrates transferability to features extracted with different networks and using different feature aggregation methods. To the best of our knowledge, little effort has been made towards automatic test-time augmentation for image retrieval. With our promising results on standard trademark and landmark datasets, we expect to attract the CV community’s attention on the potential of boosting image retrieval performance via automatic test-time augmentation.

\bibliographystyle{model2-names}
\bibliography{egbib}

\begin{thebibliography}{43}
\expandafter\ifx\csname natexlab\endcsname\relax\def\natexlab#1{#1}\fi
\providecommand{\url}[1]{\texttt{#1}}
\providecommand{\href}[2]{#2}
\providecommand{\path}[1]{#1}
\providecommand{\DOIprefix}{doi:}
\providecommand{\ArXivprefix}{arXiv:}
\providecommand{\URLprefix}{URL: }
\providecommand{\Pubmedprefix}{pmid:}
\providecommand{\doi}[1]{\href{http://dx.doi.org/#1}{\path{#1}}}
\providecommand{\Pubmed}[1]{\href{pmid:#1}{\path{#1}}}
\providecommand{\bibinfo}[2]{#2}
\ifx\xfnm\relax \def\xfnm[#1]{\unskip,\space#1}\fi
\bibitem[{Aker et~al.(2017)Aker, Tursun and Kalkan}]{aker2017analyzing}
\bibinfo{author}{Aker, C.}, \bibinfo{author}{Tursun, O.},
  \bibinfo{author}{Kalkan, S.}, \bibinfo{year}{2017}.
\newblock \bibinfo{title}{Analyzing deep features for trademark retrieval}, in:
  \bibinfo{booktitle}{Signal Processing and Communications Applications
  Conference (SIU)}.
\bibitem[{Azizpour et~al.(2015)Azizpour, Sharif~Razavian, Sullivan, Maki and
  Carlsson}]{azizpour2015generic}
\bibinfo{author}{Azizpour, H.}, \bibinfo{author}{Sharif~Razavian, A.},
  \bibinfo{author}{Sullivan, J.}, \bibinfo{author}{Maki, A.},
  \bibinfo{author}{Carlsson, S.}, \bibinfo{year}{2015}.
\newblock \bibinfo{title}{From generic to specific deep representations for
  visual recognition}, in: \bibinfo{booktitle}{Proceedings of the IEEE
  conference on computer vision and pattern recognition workshops (CVPRW)}.
\bibitem[{Babenko and Lempitsky(2015)}]{babenko2015aggregating}
\bibinfo{author}{Babenko, A.}, \bibinfo{author}{Lempitsky, V.},
  \bibinfo{year}{2015}.
\newblock \bibinfo{title}{Aggregating local deep features for image retrieval},
  in: \bibinfo{booktitle}{Proceedings of the IEEE international conference on
  computer vision (ICCV)}.
\bibitem[{Ciregan et~al.(2012)Ciregan, Meier and
  Schmidhuber}]{ciregan2012multi}
\bibinfo{author}{Ciregan, D.}, \bibinfo{author}{Meier, U.},
  \bibinfo{author}{Schmidhuber, J.}, \bibinfo{year}{2012}.
\newblock \bibinfo{title}{Multi-column deep neural networks for image
  classification}, in: \bibinfo{booktitle}{IEEE conference on computer vision
  and pattern recognition (CVPR)}.
\bibitem[{Cubuk et~al.(2019)Cubuk, Zoph, Mane, Vasudevan and
  Le}]{cubuk2019autoaugment}
\bibinfo{author}{Cubuk, E.D.}, \bibinfo{author}{Zoph, B.},
  \bibinfo{author}{Mane, D.}, \bibinfo{author}{Vasudevan, V.},
  \bibinfo{author}{Le, Q.V.}, \bibinfo{year}{2019}.
\newblock \bibinfo{title}{Autoaugment: Learning augmentation strategies from
  data}, in: \bibinfo{booktitle}{Proceedings of the IEEE conference on computer
  vision and pattern recognition (CVPR)}, pp. \bibinfo{pages}{113--123}.
\bibitem[{Dai et~al.(2017)Dai, Qi, Xiong, Li, Zhang, Hu and
  Wei}]{dai2017deformable}
\bibinfo{author}{Dai, J.}, \bibinfo{author}{Qi, H.}, \bibinfo{author}{Xiong,
  Y.}, \bibinfo{author}{Li, Y.}, \bibinfo{author}{Zhang, G.},
  \bibinfo{author}{Hu, H.}, \bibinfo{author}{Wei, Y.}, \bibinfo{year}{2017}.
\newblock \bibinfo{title}{Deformable convolutional networks}, in:
  \bibinfo{booktitle}{ICCV}.
\bibitem[{Follmann and Bottger(2018)}]{follmann2018rotationally}
\bibinfo{author}{Follmann, P.}, \bibinfo{author}{Bottger, T.},
  \bibinfo{year}{2018}.
\newblock \bibinfo{title}{A rotationally-invariant convolution module by
  feature map back-rotation}, in: \bibinfo{booktitle}{IEEE Winter Conference on
  Applications of Computer Vision (WACV)}.
\bibitem[{Gong et~al.(2014)Gong, Wang, Guo and Lazebnik}]{gong2014multi}
\bibinfo{author}{Gong, Y.}, \bibinfo{author}{Wang, L.}, \bibinfo{author}{Guo,
  R.}, \bibinfo{author}{Lazebnik, S.}, \bibinfo{year}{2014}.
\newblock \bibinfo{title}{Multi-scale orderless pooling of deep convolutional
  activation features}, in: \bibinfo{booktitle}{European conference on computer
  vision (ECCV)}.
\bibitem[{He et~al.(2015)He, Zhang, Ren and Sun}]{He2015}
\bibinfo{author}{He, K.}, \bibinfo{author}{Zhang, X.}, \bibinfo{author}{Ren,
  S.}, \bibinfo{author}{Sun, J.}, \bibinfo{year}{2015}.
\newblock \bibinfo{title}{Deep residual learning for image recognition}.
\newblock \bibinfo{journal}{arXiv preprint arXiv:1512.03385} .
\bibitem[{Hinton et~al.(2011)Hinton, Krizhevsky and
  Wang}]{hinton2011transforming}
\bibinfo{author}{Hinton, G.E.}, \bibinfo{author}{Krizhevsky, A.},
  \bibinfo{author}{Wang, S.D.}, \bibinfo{year}{2011}.
\newblock \bibinfo{title}{Transforming auto-encoders}, in:
  \bibinfo{booktitle}{International conference on artificial neural networks
  (ICANN)}.
\bibitem[{Ho et~al.(2019)Ho, Liang, Chen, Stoica and Abbeel}]{ho2019population}
\bibinfo{author}{Ho, D.}, \bibinfo{author}{Liang, E.}, \bibinfo{author}{Chen,
  X.}, \bibinfo{author}{Stoica, I.}, \bibinfo{author}{Abbeel, P.},
  \bibinfo{year}{2019}.
\newblock \bibinfo{title}{Population based augmentation: Efficient learning of
  augmentation policy schedules}, in: \bibinfo{booktitle}{International
  Conference on Machine Learning (ICML)}, pp. \bibinfo{pages}{2731--2741}.
\bibitem[{Hochreiter and Schmidhuber(1997)}]{hochreiter1997long}
\bibinfo{author}{Hochreiter, S.}, \bibinfo{author}{Schmidhuber, J.},
  \bibinfo{year}{1997}.
\newblock \bibinfo{title}{Long short-term memory}.
\newblock \bibinfo{journal}{Neural computation} .
\bibitem[{Huang et~al.(2017)Huang, Liu, van~der Maaten and
  Weinberger}]{huang2017densely}
\bibinfo{author}{Huang, G.}, \bibinfo{author}{Liu, Z.},
  \bibinfo{author}{van~der Maaten, L.}, \bibinfo{author}{Weinberger, K.Q.},
  \bibinfo{year}{2017}.
\newblock \bibinfo{title}{Densely connected convolutional networks}, in:
  \bibinfo{booktitle}{Proceedings of the IEEE Conference on Computer Vision and
  Pattern Recognition (CVPR)}.
\bibitem[{Jaderberg et~al.(2015)Jaderberg, Simonyan, Zisserman
  et~al.}]{jaderberg2015spatial}
\bibinfo{author}{Jaderberg, M.}, \bibinfo{author}{Simonyan, K.},
  \bibinfo{author}{Zisserman, A.}, et~al., \bibinfo{year}{2015}.
\newblock \bibinfo{title}{Spatial transformer networks}, in:
  \bibinfo{booktitle}{Advances in neural information processing systems
  (NeurIPS)}.
\bibitem[{Jimenez et~al.(2017)Jimenez, Alvarez and Giro-i
  Nieto}]{Jimenez_2017_BMVC}
\bibinfo{author}{Jimenez, A.}, \bibinfo{author}{Alvarez, J.M.},
  \bibinfo{author}{Giro-i Nieto, X.}, \bibinfo{year}{2017}.
\newblock \bibinfo{title}{Class-weighted convolutional features for visual
  instance search}, in: \bibinfo{booktitle}{28th British Machine Vision
  Conference (BMVC)}.
\bibitem[{Kalantidis et~al.(2016)Kalantidis, Mellina and
  Osindero}]{kalantidis2016cross}
\bibinfo{author}{Kalantidis, Y.}, \bibinfo{author}{Mellina, C.},
  \bibinfo{author}{Osindero, S.}, \bibinfo{year}{2016}.
\newblock \bibinfo{title}{Cross-dimensional weighting for aggregated deep
  convolutional features}, in: \bibinfo{booktitle}{European conference on
  computer vision (ECCV)}.
\bibitem[{Krizhevsky et~al.(2012)Krizhevsky, Sutskever and
  Hinton}]{krizhevsky2012imagenet}
\bibinfo{author}{Krizhevsky, A.}, \bibinfo{author}{Sutskever, I.},
  \bibinfo{author}{Hinton, G.E.}, \bibinfo{year}{2012}.
\newblock \bibinfo{title}{Imagenet classification with deep convolutional
  neural networks}, in: \bibinfo{booktitle}{Advances in neural information
  processing systems (NeurIPS)}.
\bibitem[{Lan et~al.(2017)Lan, Feng, Xia, Pan and Peng}]{lan2017similar}
\bibinfo{author}{Lan, T.}, \bibinfo{author}{Feng, X.}, \bibinfo{author}{Xia,
  Z.}, \bibinfo{author}{Pan, S.}, \bibinfo{author}{Peng, J.},
  \bibinfo{year}{2017}.
\newblock \bibinfo{title}{Similar trademark image retrieval integrating lbp and
  convolutional neural network}, in: \bibinfo{booktitle}{International
  Conference on Image and Graphics (ICIGP)}.
\bibitem[{Lim et~al.(2019)Lim, Kim, Kim, Kim and Kim}]{lim2019fast}
\bibinfo{author}{Lim, S.}, \bibinfo{author}{Kim, I.}, \bibinfo{author}{Kim,
  T.}, \bibinfo{author}{Kim, C.}, \bibinfo{author}{Kim, S.},
  \bibinfo{year}{2019}.
\newblock \bibinfo{title}{Fast autoaugment}, in: \bibinfo{booktitle}{Advances
  in Neural Information Processing Systems (NeurIPS)}.
\bibitem[{Marcos et~al.(2016)Marcos, Volpi and Tuia}]{marcos2016learning}
\bibinfo{author}{Marcos, D.}, \bibinfo{author}{Volpi, M.},
  \bibinfo{author}{Tuia, D.}, \bibinfo{year}{2016}.
\newblock \bibinfo{title}{Learning rotation invariant convolutional filters for
  texture classification}, in: \bibinfo{booktitle}{International Conference on
  Pattern Recognition (ICPR)}.
\bibitem[{Matsunaga et~al.(2017)Matsunaga, Hamada, Minagawa and
  Koga}]{matsunaga2017image}
\bibinfo{author}{Matsunaga, K.}, \bibinfo{author}{Hamada, A.},
  \bibinfo{author}{Minagawa, A.}, \bibinfo{author}{Koga, H.},
  \bibinfo{year}{2017}.
\newblock \bibinfo{title}{Image classification of melanoma, nevus and
  seborrheic keratosis by deep neural network ensemble}.
\newblock \bibinfo{journal}{arXiv preprint arXiv:1703.03108} .
\bibitem[{Nalepa et~al.(2019)Nalepa, Myller and Kawulok}]{nalepa2019training}
\bibinfo{author}{Nalepa, J.}, \bibinfo{author}{Myller, M.},
  \bibinfo{author}{Kawulok, M.}, \bibinfo{year}{2019}.
\newblock \bibinfo{title}{Training-and test-time data augmentation for
  hyperspectral image segmentation}.
\newblock \bibinfo{journal}{IEEE Geoscience and Remote Sensing Letters}
  \bibinfo{volume}{17}, \bibinfo{pages}{292--296}.
\bibitem[{Noh et~al.(2017)Noh, Araujo, Sim, Weyand and Han}]{noh2017large}
\bibinfo{author}{Noh, H.}, \bibinfo{author}{Araujo, A.}, \bibinfo{author}{Sim,
  J.}, \bibinfo{author}{Weyand, T.}, \bibinfo{author}{Han, B.},
  \bibinfo{year}{2017}.
\newblock \bibinfo{title}{Large-scale image retrieval with attentive deep local
  features}, in: \bibinfo{booktitle}{Proceedings of the IEEE international
  conference on computer vision (ICCV)}, pp. \bibinfo{pages}{3456--3465}.
\bibitem[{Perez et~al.(2018)Perez, Vasconcelos, Avila and
  Valle}]{perez2018data}
\bibinfo{author}{Perez, F.}, \bibinfo{author}{Vasconcelos, C.},
  \bibinfo{author}{Avila, S.}, \bibinfo{author}{Valle, E.},
  \bibinfo{year}{2018}.
\newblock \bibinfo{title}{Data augmentation for skin lesion analysis}, in:
  \bibinfo{booktitle}{OR 2.0 Context-Aware Operating Theaters, Computer
  Assisted Robotic Endoscopy, Clinical Image-Based Procedures, and Skin Image
  Analysis}. \bibinfo{publisher}{Springer}, pp. \bibinfo{pages}{303--311}.
\bibitem[{Philbin et~al.(2007)Philbin, Chum, Isard, Sivic and
  Zisserman}]{philbin2007object}
\bibinfo{author}{Philbin, J.}, \bibinfo{author}{Chum, O.},
  \bibinfo{author}{Isard, M.}, \bibinfo{author}{Sivic, J.},
  \bibinfo{author}{Zisserman, A.}, \bibinfo{year}{2007}.
\newblock \bibinfo{title}{Object retrieval with large vocabularies and fast
  spatial matching}, in: \bibinfo{booktitle}{Proceedings of the IEEE Conference
  on Computer Vision and Pattern Recognition (CVPR)}.
\bibitem[{Philbin et~al.(2008)Philbin, Chum, Isard, Sivic and
  Zisserman}]{philbin2008lost}
\bibinfo{author}{Philbin, J.}, \bibinfo{author}{Chum, O.},
  \bibinfo{author}{Isard, M.}, \bibinfo{author}{Sivic, J.},
  \bibinfo{author}{Zisserman, A.}, \bibinfo{year}{2008}.
\newblock \bibinfo{title}{Lost in quantization: Improving particular object
  retrieval in large scale image databases}, in: \bibinfo{booktitle}{2008 IEEE
  conference on computer vision and pattern recognition (CVPR)},
  \bibinfo{organization}{IEEE}. pp. \bibinfo{pages}{1--8}.
\bibitem[{Radenovi{\'c} et~al.(2018)Radenovi{\'c}, Iscen, Tolias, Avrithis and
  Chum}]{radenovic2018revisiting}
\bibinfo{author}{Radenovi{\'c}, F.}, \bibinfo{author}{Iscen, A.},
  \bibinfo{author}{Tolias, G.}, \bibinfo{author}{Avrithis, Y.},
  \bibinfo{author}{Chum, O.}, \bibinfo{year}{2018}.
\newblock \bibinfo{title}{Revisiting oxford and paris: Large-scale image
  retrieval benchmarking}, in: \bibinfo{booktitle}{Proceedings of the IEEE
  Conference on Computer Vision and Pattern Recognition}, pp.
  \bibinfo{pages}{5706--5715}.
\bibitem[{Radenovic et~al.(2018)Radenovic, Tolias and Chum}]{radenovic2018deep}
\bibinfo{author}{Radenovic, F.}, \bibinfo{author}{Tolias, G.},
  \bibinfo{author}{Chum, O.}, \bibinfo{year}{2018}.
\newblock \bibinfo{title}{Deep shape matching}, in:
  \bibinfo{booktitle}{Proceedings of the european conference on computer vision
  (eccv)}, pp. \bibinfo{pages}{751--767}.
\bibitem[{Radenovi{\'c} et~al.(2018)Radenovi{\'c}, Tolias and
  Chum}]{radenovic2018fine}
\bibinfo{author}{Radenovi{\'c}, F.}, \bibinfo{author}{Tolias, G.},
  \bibinfo{author}{Chum, O.}, \bibinfo{year}{2018}.
\newblock \bibinfo{title}{Fine-tuning cnn image retrieval with no human
  annotation}.
\newblock \bibinfo{journal}{IEEE transactions on pattern analysis and machine
  intelligence (PAMI)} .
\bibitem[{Schroff et~al.(2015)Schroff, Kalenichenko and
  Philbin}]{schroff2015facenet}
\bibinfo{author}{Schroff, F.}, \bibinfo{author}{Kalenichenko, D.},
  \bibinfo{author}{Philbin, J.}, \bibinfo{year}{2015}.
\newblock \bibinfo{title}{Facenet: A unified embedding for face recognition and
  clustering}, in: \bibinfo{booktitle}{Proceedings of the IEEE conference on
  computer vision and pattern recognition (CVPR)}, pp.
  \bibinfo{pages}{815--823}.
\bibitem[{Schulman et~al.(2017)Schulman, Wolski, Dhariwal, Radford and
  Klimov}]{schulman2017proximal}
\bibinfo{author}{Schulman, J.}, \bibinfo{author}{Wolski, F.},
  \bibinfo{author}{Dhariwal, P.}, \bibinfo{author}{Radford, A.},
  \bibinfo{author}{Klimov, O.}, \bibinfo{year}{2017}.
\newblock \bibinfo{title}{Proximal policy optimization algorithms}.
\newblock \bibinfo{journal}{arXiv preprint arXiv:1707.06347} .
\bibitem[{Seddati et~al.(2017)Seddati, Dupont, Mahmoudi and
  Parian}]{seddati2017towards}
\bibinfo{author}{Seddati, O.}, \bibinfo{author}{Dupont, S.},
  \bibinfo{author}{Mahmoudi, S.}, \bibinfo{author}{Parian, M.},
  \bibinfo{year}{2017}.
\newblock \bibinfo{title}{Towards good practices for image retrieval based on
  cnn features}, in: \bibinfo{booktitle}{Proceedings of the IEEE conference on
  computer vision and pattern recognition workshops (CVPRW)}.
\bibitem[{Sharif~Razavian et~al.(2014)Sharif~Razavian, Azizpour, Sullivan and
  Carlsson}]{sharif2014cnn}
\bibinfo{author}{Sharif~Razavian, A.}, \bibinfo{author}{Azizpour, H.},
  \bibinfo{author}{Sullivan, J.}, \bibinfo{author}{Carlsson, S.},
  \bibinfo{year}{2014}.
\newblock \bibinfo{title}{Cnn features off-the-shelf: an astounding baseline
  for recognition}, in: \bibinfo{booktitle}{Proceedings of the IEEE conference
  on computer vision and pattern recognition workshops (CVPRW)}.
\bibitem[{Simonyan and Zisserman(2014)}]{simonyan2014very}
\bibinfo{author}{Simonyan, K.}, \bibinfo{author}{Zisserman, A.},
  \bibinfo{year}{2014}.
\newblock \bibinfo{title}{Very deep convolutional networks for large-scale
  image recognition}.
\newblock \bibinfo{journal}{arXiv preprint arXiv:1409.1556} .
\bibitem[{Szegedy et~al.(2015)Szegedy, Liu, Jia, Sermanet, Reed, Anguelov,
  Erhan, Vanhoucke and Rabinovich}]{szegedy2015going}
\bibinfo{author}{Szegedy, C.}, \bibinfo{author}{Liu, W.}, \bibinfo{author}{Jia,
  Y.}, \bibinfo{author}{Sermanet, P.}, \bibinfo{author}{Reed, S.},
  \bibinfo{author}{Anguelov, D.}, \bibinfo{author}{Erhan, D.},
  \bibinfo{author}{Vanhoucke, V.}, \bibinfo{author}{Rabinovich, A.},
  \bibinfo{year}{2015}.
\newblock \bibinfo{title}{Going deeper with convolutions}, in:
  \bibinfo{booktitle}{Proceedings of the IEEE conference on computer vision and
  pattern recognition (CVPR)}, pp. \bibinfo{pages}{1--9}.
\bibitem[{Tolias et~al.(2015)Tolias, Sicre and
  J{\'e}gou}]{tolias2015particular}
\bibinfo{author}{Tolias, G.}, \bibinfo{author}{Sicre, R.},
  \bibinfo{author}{J{\'e}gou, H.}, \bibinfo{year}{2015}.
\newblock \bibinfo{title}{Particular object retrieval with integral max-pooling
  of cnn activations}.
\newblock \bibinfo{journal}{arXiv preprint arXiv:1511.05879} .
\bibitem[{Tursun et~al.(2017)Tursun, Aker and Kalkan}]{metudeeptursun}
\bibinfo{author}{Tursun, O.}, \bibinfo{author}{Aker, C.},
  \bibinfo{author}{Kalkan, S.}, \bibinfo{year}{2017}.
\newblock \bibinfo{title}{A large-scale dataset and benchmark for similar
  trademark retrieval}.
\newblock \bibinfo{journal}{CoRR} .
\bibitem[{{Tursun} et~al.(2019){Tursun}, {Denman}, {Sivapalan}, {Sridharan},
  {Fookes} and {Mau}}]{tursun2019componet}
\bibinfo{author}{{Tursun}, O.}, \bibinfo{author}{{Denman}, S.},
  \bibinfo{author}{{Sivapalan}, S.}, \bibinfo{author}{{Sridharan}, S.},
  \bibinfo{author}{{Fookes}, C.}, \bibinfo{author}{{Mau}, S.},
  \bibinfo{year}{2019}.
\newblock \bibinfo{title}{Component-based attention for large-scale trademark
  retrieval}.
\newblock \bibinfo{journal}{IEEE Transactions on Information Forensics and
  Security (TIFS)} .
\bibitem[{Tursun et~al.(2021)Tursun, Denman, Sridharan and
  Fookes}]{tursun2021learning}
\bibinfo{author}{Tursun, O.}, \bibinfo{author}{Denman, S.},
  \bibinfo{author}{Sridharan, S.}, \bibinfo{author}{Fookes, C.},
  \bibinfo{year}{2021}.
\newblock \bibinfo{title}{Learning regional attention over multi-resolution
  deep convolutional features for trademark retrieval}, in:
  \bibinfo{booktitle}{2021 IEEE International Conference on Image Processing
  (ICIP)}, \bibinfo{organization}{IEEE}.
\bibitem[{Tursun and Kalkan(2015)}]{tursun2015metu}
\bibinfo{author}{Tursun, O.}, \bibinfo{author}{Kalkan, S.},
  \bibinfo{year}{2015}.
\newblock \bibinfo{title}{Metu dataset: A big dataset for benchmarking
  trademark retrieval}, in: \bibinfo{booktitle}{2015 14th IAPR International
  Conference on Machine Vision Applications (MVA)},
  \bibinfo{organization}{IEEE}. pp. \bibinfo{pages}{514--517}.
\bibitem[{Wang et~al.(2018)Wang, Li, Aertsen, Deprest, Ourselin and
  Vercauteren}]{wang2018test}
\bibinfo{author}{Wang, G.}, \bibinfo{author}{Li, W.}, \bibinfo{author}{Aertsen,
  M.}, \bibinfo{author}{Deprest, J.}, \bibinfo{author}{Ourselin, S.},
  \bibinfo{author}{Vercauteren, T.}, \bibinfo{year}{2018}.
\newblock \bibinfo{title}{Test-time augmentation with uncertainty estimation
  for deep learning-based medical image segmentation} .
\bibitem[{Zoph and Le(2016)}]{zoph2016neural}
\bibinfo{author}{Zoph, B.}, \bibinfo{author}{Le, Q.V.}, \bibinfo{year}{2016}.
\newblock \bibinfo{title}{Neural architecture search with reinforcement
  learning}.
\newblock \bibinfo{journal}{arXiv preprint arXiv:1611.01578} .
\bibitem[{Zoph et~al.(2018)Zoph, Vasudevan, Shlens and Le}]{zoph2018learning}
\bibinfo{author}{Zoph, B.}, \bibinfo{author}{Vasudevan, V.},
  \bibinfo{author}{Shlens, J.}, \bibinfo{author}{Le, Q.V.},
  \bibinfo{year}{2018}.
\newblock \bibinfo{title}{Learning transferable architectures for scalable
  image recognition}, in: \bibinfo{booktitle}{Proceedings of the IEEE
  Conference on Computer Vision and Pattern Recognition (CVPR)}.

\end{thebibliography}

\end{document}